\documentclass[11pt,a4paper]{article}
\usepackage[hyperref]{eacl2021}
\usepackage{times}
\usepackage{latexsym}
\usepackage{graphicx}
\graphicspath{ {./images/} }
\usepackage{subcaption}
\usepackage{booktabs, multirow} 
\usepackage{soul}
\usepackage{changepage,threeparttable} 
\usepackage{tabularx}

\usepackage{microtype}

\aclfinalcopy 

\title{Does BERT Understand Sentiment? \\ Leveraging Comparisons Between Contextual and Non-Contextual Embeddings to Improve Aspect-Based Sentiment Models}

\author{Natesh Reddy \\
  Paralleldots, Inc. \\\And
  Pranaydeep Singh \\
  Paralleldots, Inc. \\
  \\
  \texttt{\{natesh,pranaydeep,muktabh\}@paralleldots.com} \\\And
  Muktabh Mayank Srivastava \\
  Paralleldots, Inc. \\
  }

\date{}

\begin{document}
\maketitle
\begin{abstract}
When performing Polarity Detection for different words in a sentence, we need to look at the words around to understand the sentiment. Massively pretrained language models like BERT can encode not only just the words in a document but also the context around the words along with them. This begs the questions, “Does a pretrain language model also automatically encode sentiment information about each word?” and “Can it be used to infer polarity towards different aspects?”. In this work we try to answer this question by showing that training a comparison of a contextual embedding from BERT and a generic word embedding can be used to infer sentiment. We also show that if we finetune a subset of weights the model built on comparison of BERT and generic word embedding, it can get state of the art results for Polarity Detection in Aspect Based Sentiment Classification datasets.
\end{abstract}

\section{Introduction}

Aspect-based Sentiment Analysis (ABSA) has always been a topic of keen research interest due to the endless commercial applications. The ability to perceive polarities associated with particular entities has several key implications for openly available social media data as well as data from websites like Amazon. To associate fine-grained sentiment with certain aspects is of a much higher degree of difficulty than standard sentence-level or document-level sentiment analysis tasks.

ABSA is divided into two sub-tasks. Firstly, Aspect extraction, the identification of entities which represent certain properties of a subject. For example in a laptop review from Amazon \textit{"The battery is fantastic but the display is a bit underwhelming"}, the words \textit{battery} and \textit{display} represent the aspects of the laptop that we want to extract. The second task, Polarity detection, aims to understand the sentiment associated with each individual aspect. In this case, the polarity for \textit{battery} would be positive, while the polarity for \textit{display} would be negative. In this work, we aim to focus primarily on the second task of Polarity detection given a set of aspects and the associated text.

Deep Neural Networks paved the way for most major developments in NLP. Even though earlier feature-based methods \cite{samha2014aspectbased} could be tailored for particular domains, DNNs generalized better. Hybrid models \cite{inbook}, combining the two approaches also yielded significant improvements. Deep Contextual Embeddings though, brought out a whole new class of models that outperformed all previous approaches. Massively pre-trained Language Models like BERT \cite{devlin2019bert} and XLNet \cite{yang2020xlnet}, yielded state-of-the-art results by simply transfer learning for downstream tasks like ABSA. Our contributions to this area are two fold:
\begin{enumerate}
  \item We demonstrate that by training to compare Deep Contextual Embeddings with standard GloVe embeddings \cite{pennington2014glove} for a particular aspect, one can obtain meaningful representations for polarity and result in a model that requires minimal compute and training time but beats baselines that take 3 times the amount of compute and training time.  
  \item We also demonstrate that when the above method is trained along with fine-tuning the last five layers of BERT, it is able to achieve state-of-the-art results on multiple standard datasets, while still being a significantly smaller and faster model when compared to recent approaches. 
\end{enumerate}

\begin{figure*}[h]
\centering
\includegraphics[scale = 0.4]{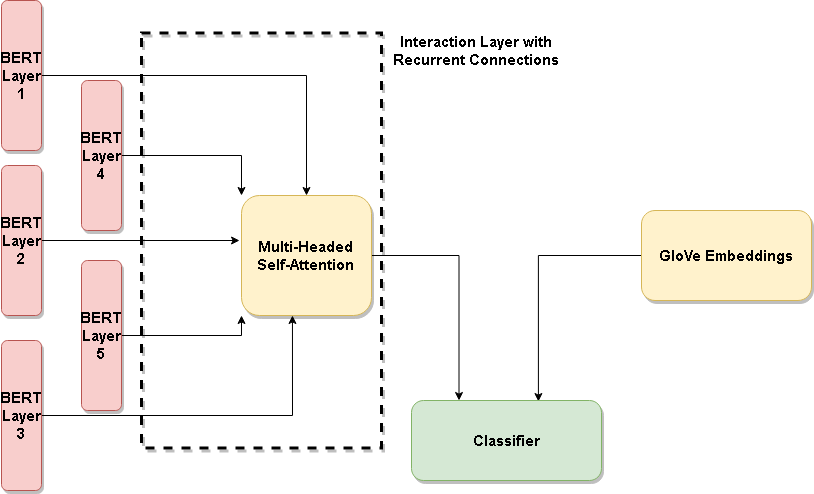} 
\caption{Simplified overview of the comparison network for BERT and GloVe Embeddings}
\end{figure*}

\section{Related Research}

While initial work in ABSA \cite{samha2014aspectbased} relied heavily on feature engineering and ontologies, deep neural networks quickly asserted their superiority. There was although a certain lack of domain specific supervision with DNNs. Hybrid approaches like HAABSA \cite{inbook} exploit domain ontologies along with deep nearal networks to get further improvements. 

With the onset of Large pre-trained Language Models, significant improvements were achieved by simply fine-tuning for downstream tasks like ABSA. Methods like BERT-QA  \cite{sun2019utilizing} further exploit the generalization abilities of BERT by framing ABSA as a Question Answering task, thus also increasing training data via augmentations for QA. Further, methods like BAT \cite{karimi2020adversarial} introduced the advantages of adversarial examples as a source of additional supervision. 

In some datasets however, methods that used domain specific ontologies were still found to be superior over neural networks with millions of parameters. BERT-PT \cite{xu2019bert}, therefore explored the advantages of domain specific pre-training for BERT embeddings to further improve domain-level context. These methods however fail to generalize on smaller domains due to the lack of pre-training data.

Our hypothesis in this work is that training to compare good contextual embeddings of aspect terms derived from BERT with generic non-contextual embeddings of the same aspects can be used to derive the sentiment the document has about the aspect terms. We use the GloVe Embedding \cite{pennington2014glove} of an aspect term as an approximation of its non-contextual embedding. GloVe Embeddings aren’t trained on domain specific data but cover data from a wide lexica. We thus assume that GloVe embeddings would provide a good enough generic non-contextual embedding. Training on a combination of BERT embeddings and GloVe embeddings is thus assumed to be able to decipher sentiment. We also explore the advantages of combining information from multiple layers of BERT using Interaction Layers like AutoInt \cite{Song_2019} to improve context resolution. We aim to achieve the above goals while keeping the trainable model size and training time very limited, in contrast to other recent approaches which extensively fine-tune BERT to increase performance.  
\begin{table*}[!htp]
\centering
\setlength\tabcolsep{3pt} 
\begin{tabular}{|l|r|r|r|r|r|r|r|r|r|r|r|r|r|}\toprule
&\multicolumn{6}{c|}{\textbf{Train}} &\multicolumn{6}{c|}{\textbf{Test}} \\\midrule
\textbf{Dataset} & \textbf{Pos} & \textbf{Neg} &\textbf{Neu} &\textbf{None} & \textbf{Conflict} &\textbf{Total} &\textbf{Pos} &\textbf{Neg} &\textbf{Neu} &\textbf{None} &\textbf{Conflict} &\textbf{Total} \\
\hline
\textbf{Semeval 14} &2176 &839 &501 &11508 &196 &15220 &657 &222 &94 &2975 &52 &4000 \\
\textbf{Semeval 16} &1400 &648 &96 &7856 &- &10000 &487 &180 &42 &2671 &- &3380 \\
\textbf{Sentihood} &1626 &834 &- &12548 &- &15008 &810 &406 &- &6300 &- &7516 \\
\bottomrule
\end{tabular}
\caption{Details of the Datasets used for the Experiments}
\end{table*}

\section{Methodology}

We build on the hypothesis that Deep Contextual Embeddings like BERT capture some semantics related to the polarity associated with the aspect, by default. This is backed by the fact that BERT or XLNet Embeddings when fine-tuned for Aspect Polarity Detection achieve near state-of-the-art results. We further hypothesize that the Polarity in the semantics can be better derived by comparing contextual GLoVE embeddings with neutral embeddings of the aspect word from standard non-Contextualized embeddings like GloVe and word2vec. To further improve the resolution of BERT Embeddings, we use output from multiple layers of BERT and aggregate them into a single vector using AutoInt, a multi-headed self-attention based interaction layer \cite{Song_2019}. 

Given a sentence \textit{s} and a set of aspects \(\{a_1, a_2, a_3 ... a_n\}\), we obtain Contextual Embeddings \(\{c_{m1}, c_{m2}, c_{m3} ... c_{mn}\}\) for each aspect from the \(m^{th}\) layer of BERT, while obtaining Non-Contextual Embeddings \(\{g_1, g_2, g_3 ... g_n\}\) from a pre-trained GloVe model. We combine embeddings from the last 5 layers of BERT using muti-headed self-attention, as this has been shown to better combine higher order features than standard weighted averaging. The modified embedding after applying the self-attention weights for a single head can be represented as,

\begin{equation}
    c'_{mn}{} = \sum_{k = 1}^{5}\alpha_{m,k}(W_{value}c_{kn})
\end{equation}

We combine outputs from 8 such heads and concatenate them, while also adding a residual connection to preserve raw features from the original BERT layers. 

\begin{equation}
    c''^{}_{mn}{} = ReLU(c'^{}_{mn}{} + Wc^{}_{mn}{})
\end{equation}

We combine the refined contextual embeddings from the 5 layers and project them to 512 dimensions along with the GloVe Embeddings \(g_{n}\) for the aspect, and then concatenate the two, followed by a Classification Layer. 

\begin{equation}
    c_{n}^{512} = W^{T}(c^{''}_{1n} \oplus c^{''}_{2n} \oplus c^{''}_{3n} \oplus c^{''}_{4n} \oplus c^{''}_{5n}) 
\end{equation}

\begin{equation}
    g_{n}^{512} = W^{T}(g_{n}) 
\end{equation}

\begin{equation}
    O_{n} = SoftMax(W^{T}(c_{n}^{512} \oplus g_{n}^{512})) 
\end{equation}

where O, is of the size \(M \times B\), where M is the number of classes and B is the batch size. We use Cross-Entropy loss, thus minimizing,

\begin{equation}
    Loss = -\sum_{c=1}^{M} Y_{c}log(O_{n,c}) 
\end{equation}

where \(Y_{c}\) is the true probability for class c, while \(O_{n,c}\) is the predicted probability for class c. 

\section{Experiments}
\label{sec:length}

\subsection{Datasets}


We carry out experiments primarily on 3 standard datasets:

\begin{itemize}
    \item SemEval 2014 Task 4 \cite{pontiki-etal-2014-semeval}: 15,220 Sentences  
    \item SemEval 2016 Task 5 \cite{pontiki-etal-2016-semeval}: 15,008 Sentences
    \item SentiHood \cite{saeidi-etal-2016-sentihood}: 10,000 Sentences
    
\end{itemize}

The details of the train-test splits and the per-label distributions are summarized in Table 1. 
 \subsection{Training}
 
 We jointly evaluate two models for all 4 of the data sets. The default model described in Section 3 is referred to as BERT-IL. The 768 dimensional BERT embeddings are obtained from the standard pre-trained BERT-base uncased model, while the 300 dimensional GloVe embeddings are obtained from standard GLoVE embeddings pre-trained on the English Wikipedia dump from 2014. We also evaluate BERT-IL Finetuned, where the final 5 layers of BERT are jointly trained alongside the BERT-IL Model. While BERT-IL Finetuned is always expected to perform better than BERT-IL, it comes at the price of a higher compute cost and memory for training and deployment. We train both models with the Adam optimizer with a learning rate of 1e-5 and Dropout of 0.1, with a batch size of 8. We report 3-way percentage accuracy scores for all the experiments.

\subsection{Results}

We compare our results with 2 baselines. The BERT baseline simply uses BERT pre-trained embeddings from the bert-base-uncased model with a classification layer. The BERT-Finetune model follows a similar approach but trains the final 5 layers of BERT, increasing the trainable parameters and also the accuracy significantly. 

\begin{table}[!htp]\centering
\scriptsize
\large
\begin{tabular}{lrr}\toprule
\textbf{BERT} & \\
\textbf{BERT-Finetune} & \\
\textbf{IMN} &83.89 \\
\textbf{BERT-PT} &84.95 \\
\textbf{BERT-SPC} &84.46 \\
\textbf{BAT} &86.01 \\
\textbf{BERT-IL} &79.5 \\
\textbf{BERT-IL Finetuned} &\textbf{86.2} \\
\bottomrule
\end{tabular}
\caption{Results for the Resturant domain data from SemEval 2014 Task 4 Subtask 2}\label{tab: }

\end{table}

Results for the Restaurant dataset from SemEval 2014 Task 4 Sub-task 2 have been summarized in Table 2. BERT-IL Finetuned achieved the best scores, improving the previous state-of-the-art of BAT by 0.2\%, while BERT-IL beats the standard BERT baseline by a significant margin with a fraction of the training time and trainable model size. We compare with other BERT based models like BERT-PT which explores a novel post training approach for RRC (Review Reading Comprehension) and ABSA. The Interactive Multi-task Learning Network (IMN) \cite{he2019interactive} explores the concept of exploiting joint training of Aspect Extraction and Polarity Detection to increase performance for both tasks. The previous state-of-the-art approach BAT leverages adversarial training along with BERT post training. While there are approaches like BERT-ADA \cite{rietzler2019adapt} that beat the scores of BAT as well, they use domain specific adaptation and additional data for pre-training which are not effective techniques for niche domains with limited data.

\begin{table}[!htp]\centering
\large

\begin{tabular}{lrr}\toprule
\textbf{BERT} & \\
\textbf{BERT-Finetune} & \\
\textbf{CABASC} &84.6 \\
\textbf{HAABSA} &88.0 \\
\textbf{HAABSA++} &87.0 \\
\textbf{BERT-IL} &80.3 \\
\textbf{BERT-IL Finetuned} &\textbf{88.7} \\
\bottomrule
\end{tabular}
\caption{Results for the SemEval 2016 Task 5 Sub-task 1 Data}\label{tab: }
\end{table}


Results for the SemEval 2016 Task 5 Sub-task 1 results are shown in Table 3. Similar to SemEval 2014, BERT-IL Finetuned achieves the best scores, beating the previous state-of-the-art of HAABSA \cite{inbook} by 0.7\%. We compare with other recent methods like HAABSA++ \cite{trusca2020hybrid} which uses Deep Contextual Embeddings along with Hierarchical Attention. Our simpler BERT-IL method here too, beats the baseline of BERT. 

\begin{table}[!htp]\centering
\large
\begin{tabular}{lrr}\toprule
\textbf{BERT} & \\
\textbf{BERT-Finetune} & \\
\textbf{Sentic-LSTM} &89.32 \\
\textbf{REN} &91.0 \\
\textbf{BERT-pair-QAM} &\textbf{93.6} \\
\textbf{BERT-IL} &85.8 \\
\textbf{BERT-IL Finetuned} &90.8 \\
\bottomrule
\end{tabular}
\caption{Results for the SentiHood dataset}\label{tab: }
\end{table}

Finally, we show results for the Sentihood dataset in Table 4. BERT-IL Finetuned seems to underperform here compared to approaches like BERT-pair-QAM \cite{sun2019utilizing} which reconstructs ABSA as a QA task. The significant difference in performance can be explained by the data augmentations introduced, which result in a dataset almost 3 times larger than vanilla Sentihood. However, here too BERT-IL bests the BERT baseline with ease.  
\begin{figure*}[h]

\begin{subfigure}{0.5\textwidth}
\includegraphics[width=0.9\linewidth, height=5cm]{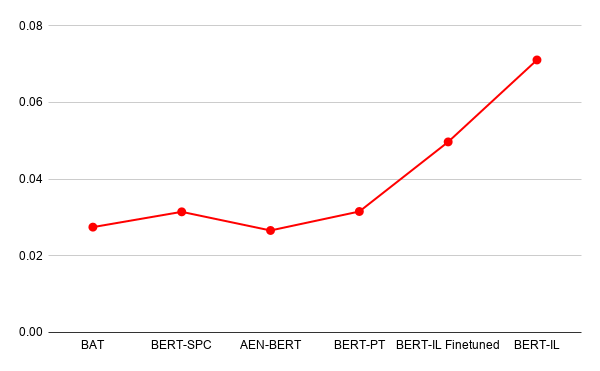} 
\caption{Accuracy/Model Size(MB) Score Comparison}
\label{fig:chart}
\end{subfigure}
\begin{subfigure}{0.5\textwidth}
\includegraphics[width=0.9\linewidth, height=5cm]{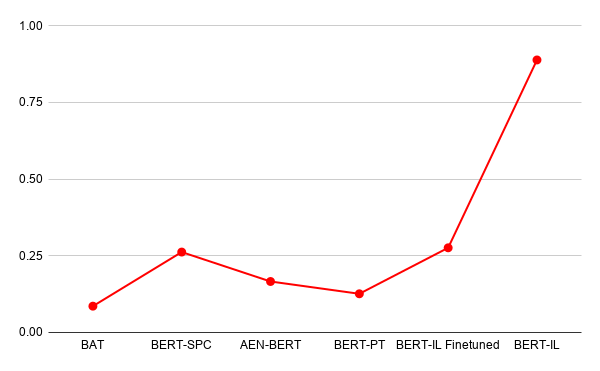}
\caption{Accuracy/Time(s) Score Comparison}
\label{fig:subim2}
\end{subfigure}

\caption{Economy and Green Scores Comparison for various recent ABSA models}
\label{fig:image2}
\end{figure*}

\begin{table*}[!htp]\centering
\begin{tabular}{lrrr}\toprule
\textbf{Model} &\textbf{Model Size (MB)} &\textbf{Time (seconds)} \\\midrule
\textbf{BAT} &{3131} &{0.32} \\
\textbf{BERT-SPC} &{2685} &{0.12} \\
\textbf{AEN-BERT} &{3125} &{0.16} \\
\textbf{BERT-PT} &{2693} &{0.25} \\
\textbf{BERT-IL} &\textbf{1119} &\textbf{0.08} \\
\textbf{BERT-IL Finetuned} &{1735} &{0.18} \\
\bottomrule
\end{tabular}
\caption{Occupied size on GPU in MBs and Time taken for a single forward and backward pass  in seconds for Batch Size 1}\label{tab: }

\end{table*}

\section{Discussion}


Deployment and training resources have been a large factor in development of the next generation of NLP models. Large pre-trained Language Models even though state-of-the-art in most downstream tasks can become taxing to even fine-tune. Not only are larger models costly to deploy, in recent times concerns of carbon emissions due to the fine-tuning of large models have also been raised. We demonstrate that with the simple BERT-IL model, without fine-tuning BERT layers, we beat baseline scores while keeping the trainable model size and training time unprecedentedly low. While with BERT-IL Finetuned we achieve state-of-the-art results on 2 standard datasets, it still manages to keep a minimal memory footprint, upto 65\% lesser than similar models, while reducing training time and thus carbon emissions by upto 75\%. 

Table 5 gives details of the Trainable Model Size(MB) and the time in seconds for a single forward and backward pass, of various models evaluated on the SemEval 14 Restaurant dataset. Since metrics like Trainable Model Size and Time for a single pass might ignore the performance of a model, we use a couple of metrics to compare the Economy and Green Scores for all the models listed. The comparisons are shown in Figure 2. 

\section{Conclusion}

We demonstrate that sentiment polarity for words in a document can be predicted by learning to compare its contextual BERT embeddings with standard Non-Contextual embeddings like GloVe. This method can be made to work just by learning a few parameters on features from pretrained models and if a few layers of the models are incrementally fine-tuned, this simple hypothesis can give competitive results. We achieve state-of-the-art results for two standard datasets with this methods, while also keeping the trainable model size and training time minimal compared to previous methods. While BERT-IL Finetuned beats previous state-of-the-art results, we also introduce the BERT-IL model which is deployment friendly and upto three times smaller and upto four times faster compared to previous methods. Because the polarity is being inferred just based on comparison of two different source of embeddings, future work on this method can also help in domain generalization of Aspect Based Sentiment Analysis tasks.



\bibliography{eacl2021}
\bibliographystyle{acl_natbib}

\end{document}